\documentclass{article}

\usepackage{arxiv}

\usepackage[utf8]{inputenc} 
\usepackage[T1]{fontenc}    
\usepackage{hyperref}       
\usepackage{url}            
\usepackage{booktabs}       
\usepackage{amsfonts}       
\usepackage{nicefrac}       
\usepackage{microtype}      
\usepackage{lipsum}
\usepackage{graphicx}
\graphicspath{ {./images/} }
\usepackage{booktabs}
\usepackage{natbib}

\title{Sociotechnical Effects of Machine Translation}

\author{
 Joss Moorkens \\
  ADAPT Centre, \\
  School of Applied Language and Intercultural Studies, \\
  Dublin City University,\\
  Dublin, Ireland \\
  \texttt{joss.moorkens@dcu.ie} \\
   \And
  Andy Way \\
  ADAPT Centre, \\
  School of Computing, \\
  Dublin City University,\\
  Dublin, Ireland \\
  \texttt{andy.way@dcu.ie} \\
  \And
 Séamus Lankford \\
 ADAPT Centre, \\
 Department of Computer Science, \\
 Munster Technological University,\\
Cork, Ireland \\
  \texttt{seamus.lankford@mtu.ie} \\
}

\begin{document}
\maketitle
\begin{abstract}
While the previous chapters have shown how machine translation (MT) can be useful, in this chapter \citep{Moorkens2024Chapter11} we discuss some of the side-effects and risks that are associated, and how they might be mitigated. With the move to neural MT and approaches using Large Language Models (LLMs), there is an associated impact on climate change, as the models built by multinational corporations are massive. They are hugely expensive to train, consume large amounts of electricity, and output huge volumes of kgCO$_{2}$ to boot. However, smaller models which still perform to a high level of quality can be built with much lower carbon footprints, and tuning pre-trained models saves on the requirement to train from scratch. We also discuss the possible detrimental effects of MT on translators and other users. The topics of copyright and ownership of data are discussed, as well as ethical considerations on data and MT use. Finally, we show how if done properly, using MT in crisis scenarios can save lives, and we provide a method of how this might be done. 
\end{abstract}


\section{Introduction}
\paragraph{}
When \citet{Weaver1949} proposed his ideas that prompted work on RBMT (as described in Chapter 1 \citep{Moorkens2024Chapter1}), he felt that even imperfect MT could contribute to “the constructive and peaceful future of the planet” by facilitating cultural interchange and international understanding. Now that imperfect, yet undoubtedly useful (and more than occasionally very good) MT is a reality, the repercussions are broader than Weaver could have imagined. We can safely say that MT facilitates cultural interchange for the over one billion users of the Google Translate Android app \citep{Pitman2021}, to name just one MT provider, but to what extent does the risk of mistranslation put those users at risk? Is MT safe to use in a crisis, such as an earthquake or pandemic? What is the relationship between MT and sustainability?
\paragraph{}
It should be quite clear by now that we are very positively disposed towards MT, but that does not mean that we are unaware of the risks of its inappropriate use. Mid-century philosophers such as \citet{Heidegger1977} tended to have a negative view of technology, but many current ethicists and philosophers of technology are more open-minded. They generally agree that technology is not necessarily good nor bad, although nor is it neutral \citep{Kranzberg1986}, but for example, \citet{Ihde1990} feels that it is the interaction with humans that is important, believing that technology “becomes what it “is” through its uses”. So what are the appropriate uses of MT? \citet{Way2013} offers a number of use-cases for MT and post-editing across a range of application areas (as well as where it should not be used) and proposes the following rule of thumb: the level of automation in translation should equate to the value, risk,
and shelf-life of a text. We can extend that to the generation of text and media using LLMs. A social media message is unlikely to be reread after a couple of hours, and an online review might be displaced by another in a similarly short time, so the shelf-life is short, and any risk arising from translation of such content is probably low. In contrast, patient instructions from a hospital or ingredients for medicines are high-risk and will need significant human oversight. A three-word slogan on a billboard is of high value, so do we really want to risk automating translation? In this case, it’s probably best not to ‘just do it’ (see \citet{Guerberof2023} for more on this topic).
\paragraph{}
The risks and repercussions of the use of MT and LLMs are broader than the decisions to use or not to use them to automatically translate (or produce, in the monolingual LLM context) a text. In this chapter \citep{Moorkens2024Chapter11}, we look at a few of these repercussions and ask the key questions laid out in Table \ref{tab:keyquestions} . The intention is not to say that MT is good or bad, but that it can have positive or negative effects through its uses. Now that you have learned how MT works and how to create your own MT system, it is important to become a critical and responsible user of MT. In the following sections, we look at some risks in the use of MT, beginning with social and environmental sustainability and MT. We come back to questions of data that we touched upon in Chapter 2 \citep{Moorkens2024Chapter2}, then consider bias, power, and diversity. Finally, we look at MT for good, such as its use in crises and show how MT, carefully deployed, can literally save lives.

\begin{table}[]
\centering
\begin{tabular}{@{}l@{}}
\toprule
\textbf{Key questions} \\ \midrule
What are the common ethical issues regarding NMT and LLMs? \\ \midrule
Are AI tools environmentally sustainable? \\ \midrule
What is the carbon footprint of an average NMT system? \\ \midrule
Are AI tools socially sustainable? \\ \midrule
What are the risks of concentrating technology and power in a small number of \\ \midrule
multinational companies? \\ \midrule
How might we apply MT for the common good? \\ \midrule
How might we empower people with NMT and LLMs? \\ \bottomrule
\\
\end{tabular}

\caption{What are the sociotechnical effects of MT?}
\label{tab:keyquestions}
\end{table}

\section{Ethical issues in MT}
\label{sec:ehtical}
\paragraph{}
The aim of this book is to empower readers with skills and knowledge for translation
automation. However, development of MT and LLMs doesn’t happen in a vacuum, and while
our small-scale systems are unlikely to have much of an effect on the outside world, it’s worth being aware of some wider context. In this final chapter we will summarise some of the ethical issues regarding MT very briefly. We have touched on these here and there earlier in the book (regarding data in Chapter 2 \citep{Moorkens2024Chapter2}, for example, and evaluation in Chapter 5 \citep{Moorkens2024Chapter5}). Publications listed in the Further Reading section will provide you with more detail and discussion, as more work emerges on these important topics as MT use continues to increase.

\subsection{MT and sustainability}
\paragraph{}
In 1987, the Brundtland Commission \citep{Brundtland1987} reported that sustainable development “meets the needs of the present without compromising the ability of future generations to meet their own needs”. With huge numbers of users, MT appears to meet the needs of current users very well, but does it compromise those future generations? In this section, we take a broad view of sustainability, beginning with environmental sustainability and then moving to social sustainability.

\subsubsection{MT and environmental sustainability}
\paragraph{}
The orthodoxy in much of the language industry and in big tech is that bigger models are always better. We showed in Chapter 2 \citep{Moorkens2024Chapter2} that NMT systems typically require amounts of data of larger magnitudes than for SMT in order to ensure good-quality output (although we’ve also seen, in Chapter 1 \citep{Moorkens2024Chapter1} and Chapter 6 \citep{Moorkens2024Chapter6}, for example, that you don’t need huge resources to build an effective system, especially in well-defined application areas). However, as  \citet{Cronin2019} writes, engagement with technology needs to be situated “within the carrying capacity of a planet with finite resources and an ever-shortening timeline of climatic viability”. With the onset of MLLMs (see Chapter 10 \citep{Moorkens2024Chapter10}), we are seeing models built with over one trillion parameters, i.e. over one trillion weights need to be optimised during training as part of an iterative process \citep{Fedus2021}. Unsurprisingly, therefore, building a model as large as GPT4 is reported to have cost over 100 million US dollars \citep{Knight2023}, a huge amount by any standards.
\paragraph{}
As well as the enormous financial cost involved, such large models (i) consume a huge amount of electricity to power the large amount of GPUs on which they are built, and (ii) output huge volumes of kgCO$_{2}$, which contributes adversely to climate change. A number of researchers have noted this and urged the NLP community as a whole to take responsibility for our role in polluting the planet, and to build smaller models with good performance that use less electricity and output fewer emissions.
\paragraph{}
\citet{Strubell2019} were among the first to attempt to quantify the environmental costs of training a variety of NLP models built using NN models. One interesting observation was that large Transformer models (see Chapter 4 \citep{Moorkens2024Chapter4}) can output more CO$_{2}$ than the lifetime of an average car, without even taking GPU and cooling requirements into account. Although these shocking figures have subsequently met with some resistance, \citet{Strubell2019} propose a number of actionable recommendations to reduce the environmental impact of the large models that we build, not least to “prioritize computationally efficient hardware and algorithms”.
\paragraph{}
One of the reasons that efficiency has not been prioritised has been a tendency to report “the single best result after running many experiments for model development and hyperparameter tuning” \citep{Schwartz2020}. Schwartz et al. suggest a mindset change to ‘Green AI’, with holistic evaluation of quality alongside efficiency. This call is taken up by \citet{Moorkens2024Triple}, who believe that “old methods that focus only on isolated measures of performance need to evolve and be enriched”. Their proposal is a ‘triple bottom line’ evaluation of people and planet alongside performance, with each receiving equal weight. The sense at present is that the first two criteria are very much an afterthought in most MT evaluation scenarios.
\paragraph{}
It appears that some MT developers (at least) are putting efforts into reducing power
requirements. \citet{Wu2021} report 25\% increases in efficiency of machine learning models over a two-year period, including equipment manufacturing and operational costs in their calculations. \citet{Jooste2022} show that NMT systems with much smaller footprints can be built which still perform well. More importantly, they demonstrate that, for an established MT service provider, these smaller models can reduce carbon emissions by almost 50\%, with a concomitant reduction in economic costs. \citet{Shterionov2023} also note that geographical location, energy provider, and time of day can make a difference in the emissions produced by NMT training.
\paragraph{}
The small-scale toolkit systems described in Chapter 6 \citep{Moorkens2024Chapter6} have low carbon footprints, plus pre-trained models save on the requirement to train from scratch (see also \citet{Dogru2024} Another advantage of using small-scale systems is that they do not require the very latest equipment, as production and disposal of technology equipment is a hidden environmental consequence, looking across the life cycle of technology \citep{Williams2011}. Nor do they have the huge cooling requirements of data centres that host systems in the cloud \citep{Townsend2023}.

\paragraph{}
A plethora of tools to evaluate the carbon footprint of NLP \citep{Bannour2021} has subsequently been developed and the concept of sustainable NLP has become an important research track in its own right at many high profile conferences such as the EACL 2021 Green and Sustainable NLP track.\footnote{https://2021.eacl.org/news/green-and-sustainable-nlp} This is very much in line with the industry trend of trying to quantify the impact of NLP on the environment.
\paragraph{}
The idea of energy awareness is now being incorporated into the development of NMT tools. In particular the adaptNMT and adaptMLLM applications, discussed in Chapters 4 \citep{Moorkens2024Chapter4} and 10 \citep{Moorkens2024Chapter10}
respectively, have a feature for tracking energy usage and kgCO$_{2}$ emissions generated during model development. Carbon emissions are calculated in a ‘green report’, primarily as an information aid, but also hopefully as a way to encourage reusable and sustainable model development. The green report embedded within the adaptNMT and adaptMLLM applications is our first implementation of a sustainable NLP feature. It is planned to develop this further to include an improved UI and user recommendations about how to develop greener models. As an open-source project, it is envisaged that the community will enthusiastically adopt these measures adding to their development by contributing new ideas and improvements.

\subsection{MT and social sustainability}
\paragraph{}
In Chapter 8 \citep{Moorkens2024Chapter8}, we noted that while post-editing continues to become more widespread, translators are reported not to like it. This seems to be less to do with MT per se than how MT is integrated into work processes. For example, \citet{Firat2024} have raised sustainability concerns due to poorly-paid post-editing jobs unilaterally imposed on translators within limited translation platforms \citep{Vieira2020}. The opportunity to use automation to cut labour costs without due consideration of the satisfaction and motivation of workers seems to be irresistible in some sectors of the market. These jobs tend to involve repetitive work on decomposed pieces of text or media, with a focus on short-term financial gains and a belief that workers are of low value and thus replaceable. The needs of workers for motivation – jobs with meaning, challenging work, a sense of responsibility and importance to an organisation – are unmet. Translation platform work also allows for more comprehensive data dispossession, with ever more data being collected to improve translation automation, and many management functions, such as job allocation, project management, and evaluation, looking to be automated. \citet{Moorkens2023Quantification} refers to this increasing requirement to satisfy algorithmic management in platform translation tasks as ‘algorithmic norms’, supplanting some of the previous norms in translation that were focused entirely on the text itself.
\paragraph{}
Consequently, there are anecdotal reports of translators leaving the language industry and claims of a talent crunch in subtitling, one of the areas of the market most affected by these practices. In contrast, \citet{Durban2022} has reiterated the need for high-quality translators at the premium end of the market. What appears to be happening is a degree of polarisation, with lower-value sections of the translation market threatened. On the other hand, technology-related employment in the language industry appears to be quite healthy, with \citet{Rothwell2023} and others highlighting the growing range of roles relating to translation for which linguists qualify. Technological skills, not least the ability to understand MT, how it works, and where and when it should be deployed, appear to be a strong differentiator. The hope is that those who successfully move into the language industry have not only these skills, but also a sense of ethics to ensure its social sustainability.
\paragraph{}
There are also social sustainability concerns regarding end users of MT, firstly that they may be exposed to unnecessary risk, particularly those with no understanding of the processes behind and risks of relying on MT, a topic now referred to as MT Literacy \cite{Bowker2019}. Secondly, there is a worry that widespread use of MT might affect languages themselves. We have seen that MT facilitates a great deal of communication, but also that MT is likely to produce some biased output and tends to have less linguistic richness and diversity than human translation \citep{Vanmassenhove2019}. As far back as 1958, \citet{Vinay1995} rather disdainfully wrote of their concern at “the prospect that four fifths of the world will have to live on nothing but translations, their intellect being starved by a diet of linguistic pap”. However, it is true that many language communities are more reliant on translation than others, and if MT is used for high-value texts, language could be at risk of being impoverished.
\paragraph{}
This is likely to persist now that MLLMs are being used for translation, albeit in a different way. In Chapter 10 \citep{Moorkens2024Chapter10}, we explained that unlike in SMT and NMT, large amounts of parallel data for the language-pair at hand do not need to be part of the training set; as long as some data for the SLs and TLs exists within the training data, the model’s inferences from whatever parallel data is included in the training set along with this monolingual SL and TL data typically ensures that reasonable quality translation for the specific language-pair in question can be achieved. However, while the TL translations look like sentences of the TL, they will have been heavily influenced by the content of the parallel data used in training, and not be reflective (at all) of the TL itself. We might effectively have a Polysystem within the MLLM in which the lesser-resourced languages are peripheral and the well-resourced central \citep{even2021position}, with a particular bias towards English that has long been recognised \citep{Bender2011}. To give an essence of this, imagine that you create a prompt for an MLLM to generate typical breakfast food, and you require the output to be in Urdu. Imagine that large amounts of English data were used to train the MLLM, and very little (if any) English--Urdu parallel data. While the MLLM may indeed generate good quality Urdu sentences which address the nature of the prompt, they are very likely to be the equivalent of ‘bacon and eggs’ in that language, which of course is not typical breakfast food in Pakistan or India, and indeed the output translations would very likely be offensive to Urdu speakers in this context. We have spoken elsewhere about the influence of the SL on translation, but this is a different kind of influence, namely that of the predominant language-pair in the training set.
\paragraph{}
The sense from \citet{Reijers2023} is that MT tends to standardise style and content, thus making it easier for commodification and exchange. A participant from \citet{Moorkens2018} makes an analogy with pre-cooked food that “always tastes the same”. Reijers and Dupont believe that this exacerbates the risks to both translators and end users, as mentioned above. This tendency to standardise is also true of texts generated by LLMs, which have been autotuned for our consumption. ChatGPT responds to our prompts in a friendly and compliant manner, with undesirable outputs avoided following the process of RLHF, the post- editing guideline to ‘edit any offensive, inappropriate or culturally unacceptable content’ effectively internalised. However, in many contexts this is actually no bad thing and fits very much with the rule of thumb from earlier in this chapter \citep{Moorkens2024Chapter11}. MT and LLM output is not ideal for every possible use case, but for lots of situations, uncontroversial and inoffensive seems like a reasonable starting point.

\subsection{Copyright and translation data}
\paragraph{}
We wrote in Chapter 2 \citep{Moorkens2024Chapter2} and elsewhere (e.g. \citet{Lewis2010}) about the grey areas around translation data and copyright. Translation is seen as a derivative work, subject to the rights of the ST author, according to the Berne convention, but while many researchers believe that there should be some rights due, particularly for translation of original or creative input, or for the creation and maintenance of databases in the form of TMs, this has not been legally tested and translation data tends to accrue in large organisations for reasons of contract or precedent. This valuable data may then become the building blocks for NMT and LLMs, alongside data from many other sources, as discussed in Chapter 2 \citep{Moorkens2024Chapter2}. 
\paragraph{}
The advent and publicity of LLMs has made this situation a little more complicated, with a broader interest in the social repercussions of AI tools and the sources of training data. Artists have discovered that their online images were used for model training, and many are using new tools that introduce minor adjustments to images, in an effort to mislead generative models \citep{Shan2023}. Thousands of authors have found that their books were among the over 191,000 used to train LLMs \citep{Reisner2023}, and the New York Times lawsuit against OpenAI alleges that millions of newspaper articles were used for training without permission \citep{Grynbaum2023}.
\paragraph{}
The European Union’s AI Act\footnote{https://www.europarl.europa.eu/topics/en/article/20230601STO93804/eu-ai-act-first-regulation-on-artificial-intelligence} looks to limit or require oversight for some high-risk uses of AI, such as job allocation (see above), but it seems unlikely that legislators -- even those behind the AI Act -- would be interested in any threats to innovation that might harm the competitiveness of companies in their regions. The decisions of judges in the many cases regarding data and copyright could nonetheless be far-reaching.

\subsubsection{Diversity in development and the rise of tech giants in MT
research}
\paragraph{}
For many years, Google Translate (GT)\footnote{https://translate.google.com/} has dominated the use of online MT (as we noted
in the introduction to this chapter \citep{Moorkens2024Chapter11}), with Bing Translator\footnote{https://www.bing.com/translator} also achieving large numbers.
Despite the existence of excellent alternative providers (especially for certain languages and language pairs, such as DeepL\footnote{https://www.deepl.com/en/translator} for German), we have become accustomed to using and relying on the engines provided by large multinational corporations (MNCs). It’s notable how the big breakthroughs tend to originate in these big MNCs with their enormous concentration of resources and power, rather than academia and smaller organisations; it is, of course, a win-win for these companies, as in order to run or tune these massive LLMs, we need to use cloud-based resources that they own! When using platforms such as OpenAI to fine-tune models, the question has to be asked - are they harvesting the data which is uploaded to tune those models? Ostensibly they don’t use your data for training their own models.\footnote{https://openai.com/policies/terms-of-use} However, if you make your custom GPT or fine-tuned model public, the “knowledge” built into the weights is now in the public domain. This can effectively be viewed as a very efficient representation of your training data. Obviously this is a concern for companies and as a consequence many firms prevent employees from using ChatGPT, Gemini, and other powerful LLM-based chatbots.
\paragraph{}
In addition, \citet{Skadina2023} note that “the multinationals who provide the services could withdraw or start charging for them at any time”. One example of a similar move in this direction was the withdrawal of rights in 2021 to prevent the use of images (or information from those images) from Google Streetview\footnote{https://www.google.com/streetview/} as a basis for other applications. Accordingly, overreliance on these tools and services always being there may be a little risky, as none of these MNCs are translation companies per se; for Google, at least, GT is a means to increase advertising revenues and maximise network effects, or take user input (perhaps from industry-internal documentation repositories) as additional future training data (see Chapter 6 \citep{Moorkens2024Chapter6} for the risks associated with using free online systems). Further evidence regarding the seriousness with which such companies take the translation problem is that Bing Translator offers translation to and from Klingon\footnote{https://www.bing.com/translator/?from=en\&to=tlh-Latn} -- a native language of precisely zero human beings -- when many other much more useful languages with huge numbers of speakers could be included instead.
\paragraph{}
Unsurprisingly, the (M)LLM approach to NLP has also been driven by large technical MNCs. As we have seen, OpenAI which built GPT4 has received huge investment from Microsoft, and all the other MNCs in the IT space have their own versions. One issue for researchers and developers who are external to the providers of these models is that they are so large that the very best we can do is fine-tune them with additional data; even here, this is a potential problem nowadays, as we can no longer guarantee that our test or tuning data is
not included in the training set when OpenAI (say) won’t tell us what’s in the latter. Note that the amount of hardware needed to train such models isn’t a problem for MNCs, many of whom are also cloud-based service providers.
\paragraph{}
We discussed the issue of the ‘space race’ towards bigger and bigger models above. In connection to their negative impact on our climate, although we noted that the trend nowadays outside of MNCs seems to be in the opposite direction, i.e. building smaller models with better performance, as it is imperative for us all to become more socially responsible. We noted in Chapter 10 \citep{Moorkens2024Chapter10} that the honeymoon period for (M)LLMs may be over given ongoing lawsuits in this space. Just before Christmas 2023, Europe brought in the AI Act following pressure to demonstrate leadership in regulating R\&D in this area; as in many areas of society, depending on the protagonists to self-regulate seems overly optimistic, so it will be interesting to see the effect on AI development in Europe, and the extent to which the rest of the world (especially the US, China and India) follow suit.

If these MNCs are allowed to continue development of ever larger models in a completely unregulated way, not only will the climate continue -- and all of us as citizens of the planet --to suffer, ultimately we all need to ask ourselves what type of future people want: do we want to be beholden to just four massive providers?\footnote{If this seems excessive, prominent commentators have also voiced their concern. In 2024 Jonathan Kanter from the U.S. Department of Justice said that “excessive concentration of power is a threat … it’s not just about prices or output but it’s about freedom, liberty and opportunity” \citep{Farrell2024}, np).} 

\section{MT for good}
\label{sec:mt4good}
\paragraph{}
There are of course, many ways to use MT for the common good beyond the topics in the
following sections, but these are intended as examples.

\subsection{MT in crisis translation}
\paragraph{}
\citet{Way2020} observe that there “have been alarmingly few attempts to provide
automatic translation services for use in crisis scenarios”. To the best of our knowledge, the first was Will Lewis’ effort \citep{Lewis2010} to build Haitian Creole systems following the devastating earthquake in 2010, as the title makes clear “from scratch in 4 days, 17 hours, \& 30 minutes”.
Estimated casualties range from 100,000 to over 300,000 deaths, with around a third of all citizens affected in some way or other by the earthquake measuring 7.0 on the Richter scale.\footnote{https://en.wikipedia.org/wiki/2010\_Haiti\_earthquake} The main issues for Lewis and his team were a complete lack of knowledge of the language (grammatical structure, encoding, orthography etc), and no data at all to train high-quality SMT engines. However, the team quickly identified some available resources (the Bible is available in most languages), and a small number of native speakers to help with translation and, especially, validation of the MT output generated. Eventually, around 150,000 segments of training data were collected to build the system, which obtained a BLEU score of almost 30 for Creole to English, and 18.3 for English to Creole, sufficiently high (especially for the into-English direction) for the system to be deployed for use by relief workers in the field. 
\paragraph{}
This remarkable effort led to the writing of a cookbook for MT in crisis scenarios \citep{Lewis2011}, so that the lessons learned from the exercise could be put into practice when other crises arose, as they do all too commonly, regrettably. Importantly, \citet{Lewis2011} note that “If done right, MT can dramatically increase the speed by which relief can be provided”. In any such scenario, translation is almost always needed, and despite its importance, it is often overlooked. 
\paragraph{}
In response to the need for better preparation for translation readiness in crises, our colleague Sharon O’Brien coordinated the Interact project.\footnote{https://sites.google.com/view/crisistranslation/home} \citep{Federici2019} provides a set of recommendations within that project which apply mainly to human translation provision in crisis scenarios.
\paragraph{}
One recent crisis that affected us all is still around today. China was the first country to report outbreaks of the COVID-19 virus in late 2019. A month later, the World Health Organization declared the outbreak to be an international public health emergency, and on 11th March 2020, they upgraded it to a pandemic.\footnote{https://www.who.int/emergencies/diseases/novel-coronavirus-2019/events-as-they-happen} As we know now, different countries reacted differently to try to curb the spread of the virus, but as the virus was airborne, its effects transcended national borders, and an international coordinated response was needed to limit the damage.
\paragraph{}
As certain countries bore the brunt of the repercussions of the disease at different times, it was important to learn from those affected early on, which necessitated the translation of information from one language to another. One very early attempt to level the playing field in this regard was the building of eight high-performing MT systems tuned to this domain by \citet{Way2020}. Importantly, these engines had free access to all, thus “empowering individuals to access multilingual information that otherwise might be denied them”. The systems were compared against freely available MT systems, and the performance of the engines was comparable, meaning that users could translate material available in French, Italian, German and Spanish (sometimes known as FIGS) to and from English, with none of the usual security concerns (see Chapter 6 \citep{Moorkens2024Chapter6}) associated with online systems. An additional advantage of self-build was that \citet{Way2020} maintained control of the engines, allowing them to be continuously improved following the receipt of user feedback.
\paragraph{}
\citet{Way2020} adapt the recommendations in \citet{Federici2019} to MT, noting the potential to help (numbered recommendations are those used by \citet{Federici2019}):
\begin{itemize}
\item “improve response, recovery and risk mitigation [by including] mechanisms to provide accurate translation” [Recommendation 1a, 8]
\item “address the needs of those with heightened vulnerabilities [such as] … the elderly” 
[Recommendation 1b, 9]
\item those “responsible for actioning, revising and training to implement … translation policy within …organization[s]” [Recommendation 2a, 9]
\end{itemize}

\paragraph{}
Recommendation 7a \citep{Federici2019} notes that “Translating in one direction is insufficient. Two-way translated communication is essential for meeting the needs of crisis and disaster-affected communities.” \citet{Way2020} built engines for both directions (FIGS-to-English as well as English-to-FIGS), thus facilitating two-way communication, which would be essential in a patient-carer situation in a crisis scenario, among many others. Two benefits noted by \citet{Way2020} of building MT engines to do the translation instead of using human professionals included the avoidance of (i) needing to train translators, and (ii) exposing human translators to traumatic situations (Recommendation 8d: \citep{Federici2019}). 
\paragraph{}
This work on building MT systems for COVID-related documentation was extended by \citet{Lankford2021} for the Irish language. In developing MT systems during a crisis, there are many considerations to be borne in mind. Central to these considerations is how much time is available to provide a solution which provides value and is also acceptable to stakeholders. With this in mind, it is worthwhile to consider a range of options for delivering MT solutions.
\paragraph{}
As with any crisis, quality may initially suffer at the sacrificial altar of expediency, so the following three-pronged approach is one solution which practitioners may choose to adopt. The approach delivers a rapid-response lower-quality MT system which subsequently evolves to a higher-quality model. For a more detailed description on how to implement each of these steps, we refer readers to Chapter 10 \citep{Moorkens2024Chapter10}.

\subsection{How might we deploy MT in crisis translation?}
\paragraph{}
Our approach to enhancing MT in crisis situations involves three key strategies. Initially, a custom GPT is created on the ChatGPT platform\footnote{https://chat.openai.com/gpts} immediately after a crisis, enabling users to contribute to a specialised knowledge base with new terms relevant to the crisis, effectively crowdsourcing a dataset for crisis-specific language pairs. As time permits, a new GPT should be fine-tuned within the OpenAI platform to develop a model with new weights tailored to the specific language needs of the crisis.
\paragraph{}
Finally, a bespoke model can be created using an open-source tool like adaptMLLM, fine-tuned with a custom dataset developed during the crisis, similar perhaps to what \citet{Lewis2010} did. Such a phased approach allows for rapid initial response and progressively more tailored MT solutions as the crisis unfolds, leveraging community input and specialised training to
improve translation accuracy in critical situations.

\begin{figure}[ht]
    \centering
    \includegraphics[scale=0.8]{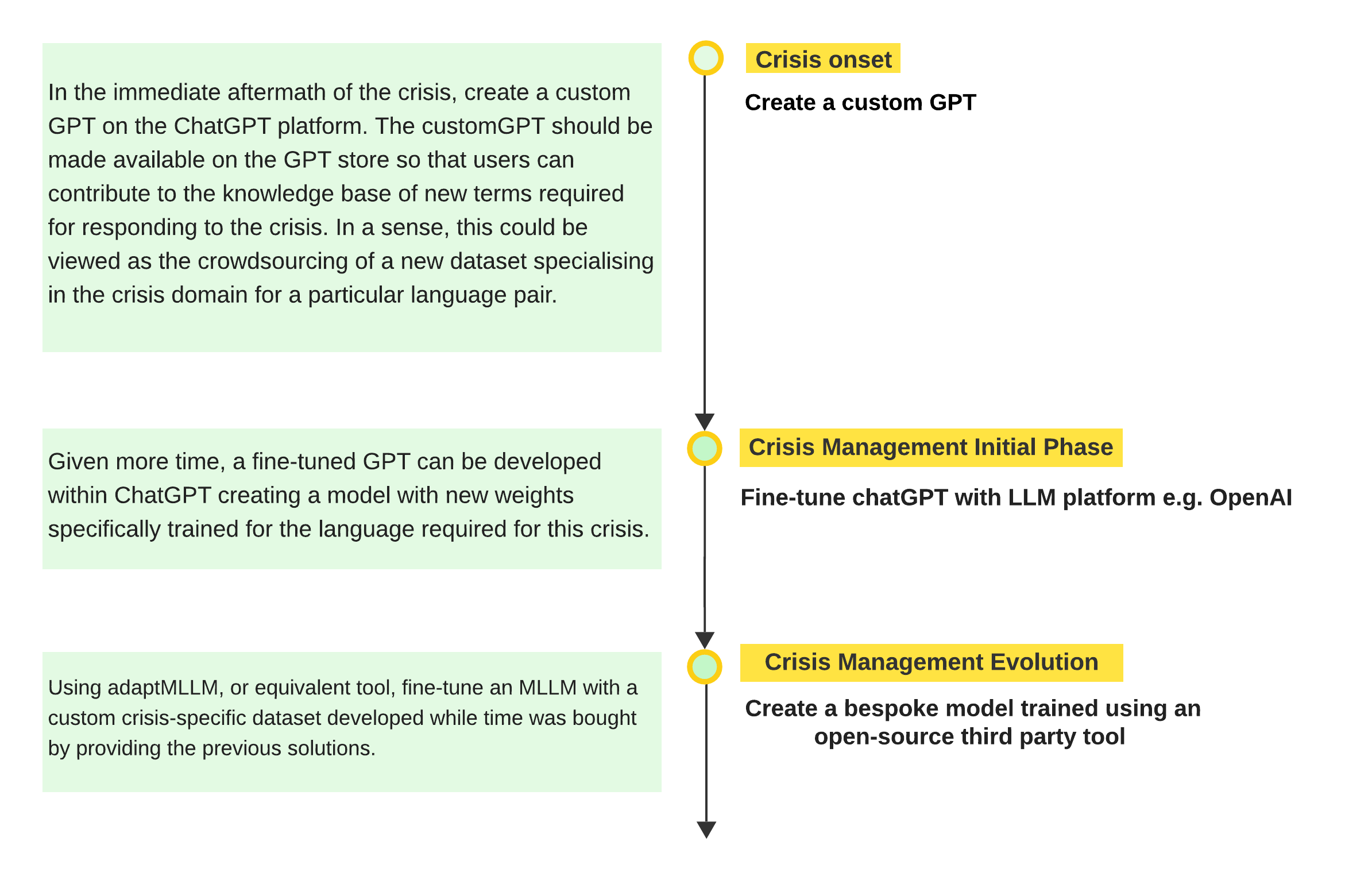}
    \caption{From expediency to quality: how to rapidly deploy LLMs for crisis MT.}
    \label{fig:crisisline}
\end{figure}

Of course, a major consideration when designing an MT system in crisis scenarios is the availability of a suitable parallel corpus which contains new terminology associated with the unfolding crisis. However, it is precisely at these times when the production of such datasets presents the greatest challenge as we showed above. 
\paragraph{}
As part of our research work in developing a bespoke parallel health corpus for the Irish language specialising in the COVID domain, a set of guidelines was developed which formulates a process for dataset development. This approach is particularly suitable for low-resource languages.
\paragraph{}
PDF and Word documents were pre-processed using a toolchain currently under development as part of the Irish Language Resource Infrastructure project (ILRI),\footnote{https://www.adaptcentre.ie/projects/european-language-resource-infrastructure/} funded by the Department of the Gaeltacht within the Irish government. This toolchain has been written to primarily accept data that originates in public administration organisations, i.e. relatively formal text for which the translation quality is assumed to be high, the structure/formatting to be reasonably consistent, and the potential for noise to be low.
\paragraph{}
PDFs in particular can be problematic for creating high-quality corpora for a variety of reasons (see \citet{Poncelas2020} for some of these). In other words, while the quality of the input content in this case can be said to be high, the quality of the input medium is low. The process used for developing the corpus is illustrated in Figure \ref{fig:crisisline}. The toolchain consists of a set of
components run in sequence over a set of input documents, in order to convert them from raw content to a sentence-aligned corpus. Several of the components listed here have different implementations depending on the source type and intended output.
\paragraph{}
With the above considerations in mind, the set of rules laid out in "Guidelines for standardising parallel corpora" was decided upon when processing the gaHealth corpus.\footnote{https://github.com/seamusl/gaHealth} Many of these could be specified as parameters to the toolchain, while others were hard coded into the system.

\paragraph{Guidelines for standardising parallel corpora}

\begin{enumerate}

\item Unicode standard: normalise all characters to Unicode UTF-8 NFC (see Chapter 2 \citep{Moorkens2024Chapter2}for a discussion on these formats). Remove any byte order marks.
\item Whitespacing and capitalisation: merge sequences of whitespace characters into a single space. Do not perform tokenisation or truecasing.
\item File language detection: scan the first 50 lines, and then every 100th line.
\item Document alignment: assume that specific patterns like a line beginning with a single letter in parentheses or a number followed by a full stop indicate a sentence break from the previous line. Ensure each document is 0.75-1.33 times the size of the document it is being aligned with. Run for a maximum of three iterations.
\item Sentence alignment: allow one-to-many alignments.
\item Cleaning: remove any pairs where source or target:
\begin{enumerate}
  \item is empty
  \item contains no non-alphabetical characters
  \item is of an incorrect language. This will remove most untranslated segments.The language is only to be detected for segments that have at least 40 characters.
\end{enumerate}
\end{enumerate}

\begin{figure}[ht] 
    \centering
    \includegraphics[scale=0.7]{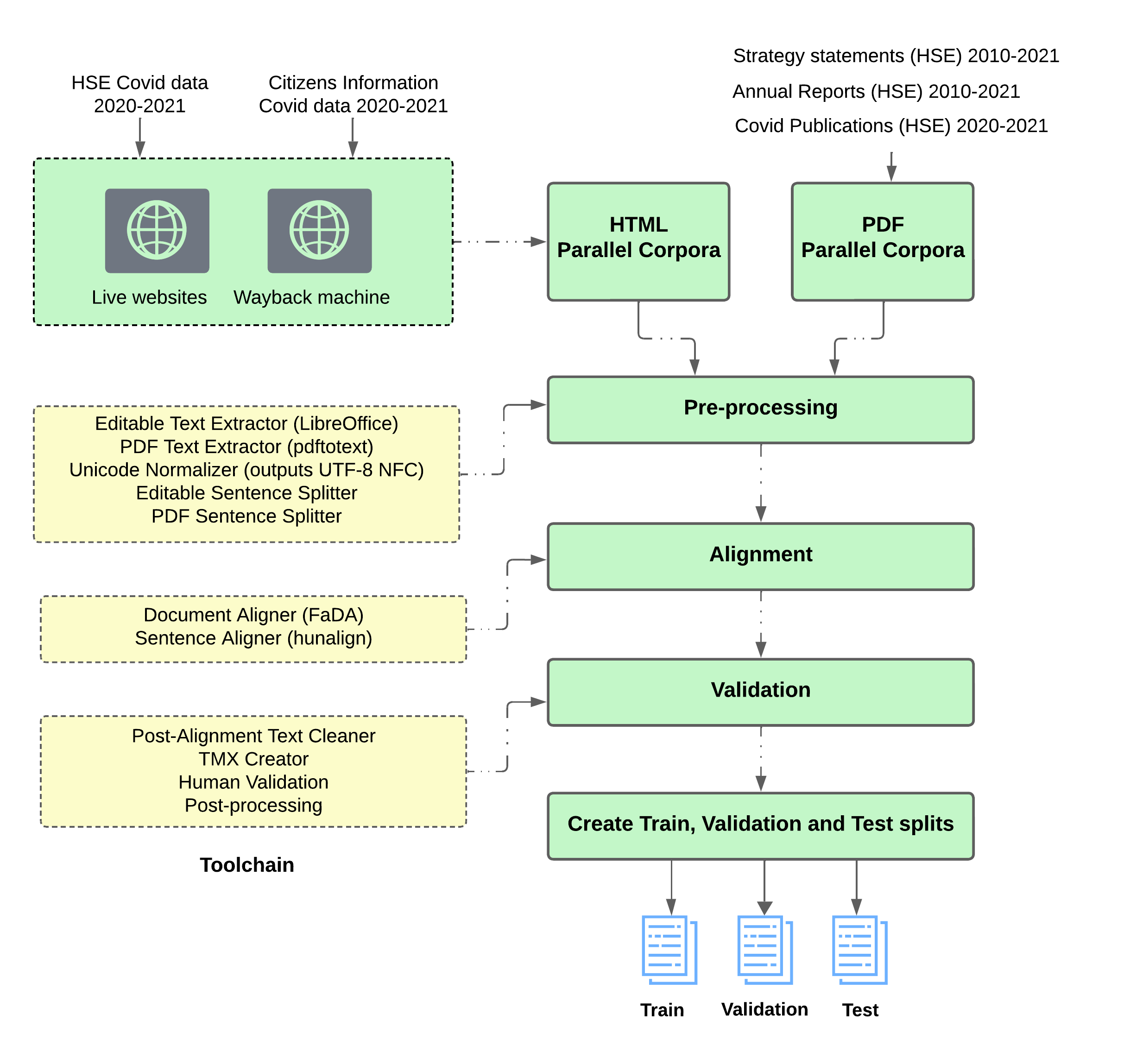}
    \caption{Corpus development process.}
  \label{fig:corpusdev}
\end{figure}

\paragraph{}
In developing the corpus, the key steps of data collection, pre-processing, alignment and validation were followed. The role of the toolchain at various stages is highlighted in Figure \ref{fig:corpusdev}. A detailed description of the individual components within the toolchain is available in \citet{Lankford2022}.

\section{AI and automation for empowerment}
\label{sec:aiauto}
\paragraph{}
Throughout the book we have used the term AI, not because we believe LLMs to be intelligent or because we’re fond of the term, but rather because it’s widely used for these technologies as more or less synonymous with machine learning. However, AI predates this, and was used for knowledge-based expert systems in a similar manner to RBMT. The common definitions of AI and automation tend to – rather unhelpfully – pit humans against machines. Thus, AI is often defined as thinking or behaving like humans, or as behaving rationally, and automation as a “device or system that accomplishes (partially or fully) a function that was previously, or conceivably could be, carried out (partially or fully) by a human operator” \citep{Parasuraman2000}. \citet{OBrien2023} calls these oppositional definitions ‘antagonistic dualisms’, and asks why we can’t envisage AI not as a replacement for humans, but rather as a way to work in combination to achieve what was previously not possible.
\paragraph{}
\citet{OBrien2023} cites work by \citet{Shneiderman2020}, who believes that humans can simultaneously have more control alongside more automation, quite the opposite of what we see in every typology of automation (see \citet{Vagia2016}), and that “the approach should most definitely not be emulation, but application”. This will be challenging, as applications are variable, and the augmentation that O’Brien suggests requires feedback on cognitive states in order for the machine to best support the human.
\paragraph{}  
It’s clear that MT currently helps an awful lot of people. The work of \citet{Vollmer2020} and \citet{Ciribuco2020} shows that vulnerable users (migrants and asylum seekers) can be empowered by MT and can acquire MT literacy, but this also shows that they have to learn to work around the machine. The work of \citet{Liebling2020} with immigrants in the United States also shows that vulnerable users can be disempowered and embarrassed when the MT system produces errors that they reproduce or disseminate. The arguments about AI alignment – aligning the systems with human values – still attribute agency to the machine, which is encoded with principles and objectives (although whose principles are unclear; \citet{Gabriel2020}). As \citet{Meadows2008} wrote, “how do we optimise; we don’t even know what to optimise”.
\paragraph{}
Another question is: How do we align with a dynamic world? Our current AI technologies, such as NMT and LLMs, are moments in history frozen in time from the point at which data was collected. Our judgement on the risks of using MT and LLMs is also historical, "based on a certain programmed past" \citep{Reijers2023}. The black-box nature of NMT and LLMs means that they are inherently unpredictable, with hallucinations (output untethered from the input) and emergent properties as LLMs scale up \citep{Wei2022}. Whether or not the augmentation that O’Brien describes is possible, the aim of using AI and automation to enable what couldn’t be done before seems like an admirable direction of travel. The challenge, again citing \citep{Meadows2008} is to build a system so that “its properties and our values can work together to bring forth something much better than could ever be produced by our will alone”.

\section{Follow-on tasks and reflection}
\label{sec:followon}
\paragraph{}
By the very nature of crises, they are relatively unpredictable, but some knowledge of global politics can give pointers to the sorts of things that are likely to happen. For example, while not a full-blown crisis per se, the 2024 volcanic eruption in Iceland was foreseen, and Grindavík, the town most likely to be impacted by the eruption, was evacuated with no loss of life. The earthquake in Petrinja in Croatia in December 2020, as well as the earlier one
that year in the capital Zagreb -- in the middle of the COVID-19 pandemic, note -- did result in fatalities, but previous risk assessment \citet{Atalic2021} was vital in ensuring an appropriate response.
\paragraph{}
While neither Icelandic nor Croatian are well-resourced languages, they are countries where English-language competence is high, so even if foreign aid workers were needed, communication is unlikely to have been hugely impaired. Nevertheless, as we have seen, emergencies escalate into crises when they occur in areas where major international languages are not widely spoken.
\paragraph{}
Accordingly, imagine a particular crisis emerging in a region where the languages spoken are resource-poor. Imagine that crisis experts are flown in from countries where you speak the dominant language, and when they arrive, communication is almost impossible with the local population most impacted by the disaster. What resources could you think of which might be useful for seeding an MT system that might improve communication? These could be monolingual or multilingual resources. Which ones are easier to acquire quickly? How could each resource type be used to quickly build an MT system? What type of MT system would you suggest might be built most quickly? What quality assurance techniques could you think of to indicate that the system was likely to be effective? For what type of data should that MT system be used? Are there instances where the MT system ought not to be used?

\section{Further reading}
\label{sec:further}
\paragraph{}

For an introduction to the topic, Translation Ethics by \citet{Lambert2023} in this series is a good place to begin. Another very comprehensive book is the Routledge Handbook of Translation and Ethics, edited by \citet{Koskinen2021}, which includes a chapter on ethics in the translation industry by Moorkens and Rocchi.

\paragraph{}
For work on ethics and the sociotechnical aspects of MT, the special issue of Translation Spaces on Fair MT, with an introduction by \citet{Kenny2020} looks at various topics that we’ve touched on in this chapter \citep{Moorkens2024Chapter11}. “Towards Responsible Machine Translation”, edited by \citet{Moniz2023}, is another good choice. The chapter on ethics and MT by \citet{Moorkens2022} from Machine Translation for Everyone provides a short and readable introduction.
 
\paragraph{}
For readers interested in reading more about crisis translation and MT, a very good place to start is the list of publications at the Interact project site.\footnote{https://sites.google.com/view/crisistranslation/publications}Another project on predicting refugee migration in crisis scenarios coordinated by our colleague Haithem Afli is the ITFlows project\footnote{https://www.itflows.eu/}, which has a list of available
documents in this critical area.
\paragraph{}
Finally, some other relevant articles are \citet{Moorkens2023Quantification} on quantification and algorithmic norms in translation and \citet{Moorkens2024Triple} on proposing a sustainable ‘triple bottom line’ of people, planet, and performance for translation evaluation. 


\bibliographystyle{plainnat}  
\bibliography{references}

\end{document}